\RequirePackage{fix-cm}
\documentclass[smallextended]{svjour3}       
\smartqed  
\usepackage{graphicx,url}
\usepackage{xcolor,colortbl}
\usepackage[cmex10]{amsmath}

\journalname{Neural Processing Letters}

\begin{document}

\title{Temperature-Based Deep Boltzmann Machines}

\author{Leandro Aparecido Passos J\'unior         \and
        Jo\~ao Paulo Papa 
}


\institute{L. Passos \at
              Department of Computing, Federal University of S\~ao Carlos \\
              Tel.: +55-16-3351-8232 / Fax: +55-16-3351-8233\\
              \email{leandropassosjr@gmail.com}
           \and
           J. Papa \at
              Department of Computing, S\~ao Paulo State University\\
              Tel./Fax: +55-14-3103-6079\\
              \email{papa@fc.unesp.br}
}

\date{Received: date / Accepted: date}

\maketitle

\begin{abstract}
Deep learning techniques have been paramount in the last years, mainly due to their outstanding results in a number of applications, that range from speech recognition to face-based user identification. Despite other techniques employed for such purposes, Deep Boltzmann Machines are among the most used ones, which are composed of layers of Restricted Boltzmann Machines (RBMs) stacked on top of each other. In this work, we evaluate the concept of temperature in DBMs, which play a key role in Boltzmann-related distributions, but it has never been considered in this context up to date. Therefore, the main contribution of this paper is to take into account this information and to evaluate its influence in DBMs considering the task of binary image reconstruction. We expect this work can foster future research considering the usage of different temperatures during learning in DBMs.
\end{abstract}

\section{Introduction}
\label{introduction.tex}

Deep learning techniques have attracted considerable attention in the last years due to their outstanding results in a number of applications~\cite{GohECCV:12,DuongCVPR:15,SohnNIPS:15}, since such techniques possess an intrinsic ability to learn different information at each level of a hierarchy of layers~\cite{Lecun15}. Restricted Boltzmann Machines (RBMs)~\cite{Hinton:12}, for instance, are among the most pursued techniques, even though they are not deep learning-oriented themselves, but by building blocks composed of stacked RBMs on top of each other one can obtain the so-called Deep Belief Networks (DBNs)~\cite{Hinton:06} or the Deep Boltzmann Machines (DBMs)~\cite{Salakhutdinov:12}, which basically differ from each other by the way the inner layers interact among themselves.

The Restricted Boltzmann Machine is a probabilistic model that uses a layer of hidden units to model the distribution over a set of inputs, thus compounding a generative stochastic neural network~\cite{larochelle2012learning,schmidhuber2015deep}. RBMs were firstly idealized under the name of ``Harmonium"\ by Smolensky in 1986~\cite{smolensky1986information}, 
and some years later renamed to RBM by Hinton et. al.~\cite{Hinton:02}. Since then, the scientific community has been putting a lot of effort in order to improve the results in a number of application that somehow make use of RBM-based models~\cite{HinSal06,hinton2011discovering,PapaGECCO:15,PapaJoCS:15,PapaASC:15,TomczakNPL:16}.

Roughly speaking, the key role in RBMs concerns their learning parameter step, which is usually carried out by sampling in Markov chains in order to approximate the gradient of the logarithm of the likelihood concerning the estimated data with respect to the input one. In this context, Li et. al.~\cite{li2016temperature} recently highlighted the importance of a crucial concept in Boltzmann-related distributions: their ``temperature", which has a main role in the field of statistical mechanics~\cite{mendes2015nonlinear},~\cite{e2014statistical},~\cite{gadjiev2015origin}, idealized by Wolfgang Boltzmann. In fact, a Maxwell-Boltzmann distribution~\cite{gordon2002maxwell,shim2010robust,niven2005exact} is a probability distribution of particles over various possible energy states without interacting with one another, expect for some very brief collisions, where they exchange energy. Li et. al.~\cite{li2016temperature} demonstrated the temperature influences on the way RBMs fire neurons, as well as they showed its analogy to the state of particles
in a physical system, where a lower temperature leads to a lower particle activity, but higher entropy~\cite{bekenstein1973black},~\cite{rrnyi1961measures}.

However, as far we are concerned, the impact of different temperatures during the Markov sampling has never been considered in Deep Boltzmann Machines. Therefore, the main contributions of this work are two fold: (i) to foster the scientific literature regarding DBMs, and (ii) to evaluate the impact of temperature during DBM learning phase. Also, we considered Deep Belief Networks for comparison purposes concerning the task of binary image reconstruction over three public datasets. The remainder of this paper is organized as follows. Section~\ref{s.dbm} presents the theoretical background related to DBMs and the proposed temperature-based approach, and Section~\ref{s.methodology} describes the methodology adopted in this work. The experimental results are discussed in Section~\ref{s.experiments}, and conclusions and future works are stated in Section~\ref{s.conclusions}.

\section{Deep Boltzmann Machines}
\label{s.dbm}

In this section, we briefly explain the theoretical background related to RBMs and DBMs.

\subsection{Restricted Boltzmann Machines}
\label{ss.rbm}

Restricted Boltzmann Machines are energy-based stochastic neural networks composed of two layers of neurons (visible and hidden), in which the learning phase is conducted by means of an unsupervised fashion. 
A na\"ive architecture of a Restricted Boltzmann Machine comprises a visible layer $\textbf{v}$ with $m$ units and a hidden layer $\textbf{h}$ with $n$ units. Additionally, a real-valued matrix $\textbf{W}_{m\times n}$ models the weights between the visible and hidden neurons, where $w_{ij}$ stands for the weight between the visible unit $v_i$ and the hidden unit $h_j$.


Let us assume both $\textbf{v}$ and $\textbf{h}$ as being binary-valued units. In other words, $\textbf{v}\in\{0,1\}^m$ e $\textbf{h}\in\{0,1\}^n$. The energy function of a Restricted Boltzmann Machine is given by:

\begin{equation}
\label{e.energy_bbrbm}
E(\textbf{v},\textbf{h})=-\sum_{i=1}^ma_iv_i-\sum_{j=1}^nb_jh_j-\sum_{i=1}^m\sum_{j=1}^nv_ih_jw_{ij},
\end{equation}
where $\textbf{a}$ e $\textbf{b}$ stand for the biases of visible and hidden units, respectively.

The probability of a joint configuration $(\textbf{v},\textbf{h})$  is computed as follows:

\begin{equation}
\label{e.probability_configuration}
P(\textbf{v},\textbf{h})=\frac{1}{Z}e^{-E(\textbf{v},\textbf{h})},
\end{equation}
where $Z $ stands for the so-called partition function, which is basically a normalization factor computed over all possible configurations involving
the visible and hidden units. Similarly, the marginal probability of a visible (input) vector is given by:

\begin{equation}
P(\textbf{v})=\frac{1}{Z}\displaystyle\sum_{\textbf{h}}e^{-E(\textbf{v},\textbf{h})}.
\end{equation}

Since the RBM is a bipartite graph, the activations of both visible and hidden units are mutually independent, thus leading to the following conditional probabilities:

\begin{equation}
	P(\textbf{v}|\textbf{h})=\prod_{i=1}^mP(v_i|\textbf{h}),
\end{equation}
and

\begin{equation}
	P(\textbf{h}|\textbf{v})=\prod_{j=1}^nP(h_j|\textbf{v}),
\end{equation}
where

\begin{equation}
\label{e.probv}
P(v_i=1|\textbf{h})=\phi\left(\sum_{j=1}^nw_{ij}h_j+a_i\right),
\end{equation}

and

\begin{equation}
\label{e.probh}
P(h_j=1|\textbf{v})=\phi\left(\sum_{i=1}^mw_{ij}v_i+b_j\right).
\end{equation}
Note that $\phi(\cdot)$ stands for the logistic-sigmoid function.

Let $\theta=(W,a,b)$ be the set of parameters of an RBM, which can be learned through a training algorithm that aims at maximizing the product of probabilities given all the available training data ${\cal V}$, as follows:

\begin{equation}
	\arg\max_{\Theta}\prod_{\textbf{v}\in{\cal V}}P(\textbf{v}).
\end{equation}
One can solve the aforementioned equation using the following derivatives over the matrix of weights \textbf{W}, and biases $\textbf{a}$ and $\textbf{b}$ at iteration $t$ as follows:

\begin{equation}
\label{e.updateW2}
\textbf{W}^{t+1}=\textbf{W}^t+\underbrace{\eta(P(\textbf{h}|\textbf{v})\textbf{v}^T-P(\tilde{\textbf{h}}|\tilde{\textbf{v}})\tilde{\textbf{v}}^T)+\Phi}_{=\Delta\textbf{W}^t},
\end{equation}

\begin{equation}
\label{e.updatea2}
\textbf{a}^{t+1}=\textbf{a}^t+\underbrace{\eta(\textbf{v}-\tilde{\textbf{v}})+\alpha\Delta \textbf{a}^{t-1}}_{=\Delta\textbf{a}^t}
\end{equation}
and

\begin{equation}
\label{e.updateb2}
\textbf{b}^{t+1}=\textbf{b}^t+\underbrace{\eta(P(\textbf{h}|\textbf{v})-P(\tilde{\textbf{h}}|\tilde{\textbf{v}}))+\alpha\Delta \textbf{b}^{t-1}}_{=\Delta\textbf{b}^t},
\end{equation}
where $\eta$ stands for the learning rate, and $\lambda$ and $\alpha$ denote the weight decay and the momentum, respectively. Notice the terms $P(\tilde{\textbf{h}}|\tilde{\textbf{v}})$ and $\tilde{\textbf{v}}$ can be obtained by means of the Contrastive Divergence~\cite{Hinton:02} technique, which basically ends up performing Gibbs sampling using the training data as the visible units. Roughly speaking, Equations~\ref{e.updateW2},~\ref{e.updatea2} and~\ref{e.updateb2} employ the well-known Gradient Descent as the optimization algorithm. The additional term $\Phi$ in Equation~\ref{e.updateW2} is used to control the values of matrix $\textbf{W}$ during the convergence process, and it is formulated as follows:

\begin{equation}
\label{e.theta}
	\Phi = -\lambda\textbf{W}^t+\alpha\Delta\textbf{W}^{t-1}.
\end{equation}

\subsection{Deep Boltzmann Machines}
\label{ss.dbm}

Learning more complex and internal representations of the data can be accomplished by using stacked RBMs, such as DBNs and DBMs. In this paper, we are interested in the DBM formulation, which 
is slightly different from DBN one. Suppose we have a DBM with two layers, where $\textbf{h}^1$ and $\textbf{h}^2$ stand for the hidden units at the first and second layer, respectively. 

The energy of a DBM can be computed as follows:

\begin{equation}
	\label{e.dbm_energy}
	E(\textbf{v},\textbf{h}^1,\textbf{h}^2)=-\sum_{i=1}^{m^1}\sum_{j=1}^{n^1}v_ih^1_jw^1_{ij}-\sum_{i=1}^{m^2}\sum_{j=1}^{n^2}h^1_ih^2_jw^2_{ij},
\end{equation}
where $m^1$ and $m^2$ stand for the number of visible units in the first and second layers, respectively, and $n^1$ and $n^2$ stand for the number of hidden units in the first and second layers, 
respectively. In addition, we have the weight matrices $\textbf{W}^1_{m^1\times n^1}$ and $\textbf{W}^2_{m^2\times n^2}$, which encode the weights of the connections between vectors $\textbf{v}$ and 
$\textbf{h}^1$, and vectors $\textbf{h}^1$ and $\textbf{h}^2$, respectively. For the sake of simplification, we dropped the bias terms out.

The marginal probability the model assigns to a given input vector $\textbf{v}$ is given by:

\begin{equation}
\label{e.probability_sample_dbm}
P(\textbf{v})=\frac{1}{Z}\displaystyle\sum_{\textbf{h}^1,\textbf{h}^2}e^{-E(\textbf{v},\textbf{h}^1,\textbf{h}^2)}.
\end{equation}
Finally, the conditional probabilities over the visible and the two hidden units are given as follows:

\begin{equation}
\label{e.probv_dbm}
	P(v_i=1|\textbf{h}^1)=\phi\left(\sum_{j=1}^{n^1}w^1_{ij}h^1_j\right),
\end{equation}

\begin{equation}
\label{e.probh2_dbm}
	P(h^2_z=1|\textbf{h}^1)=\phi\left(\sum_{i=1}^{m^2}w^2_{iz}h^1_i\right),
\end{equation}
and 

\begin{equation}
\label{e.probh1_dbm}
	P(h^1_j=1|\textbf{v},\textbf{h}^2)=\phi\left(\sum_{i=1}^{m^1}w^1_{ij}v_i+\sum_{z=1}^{n^2}w^2_{jz}h^2_z\right).
\end{equation}

After learning the first RBM using Contrastive Divergence, for instance, the generative model can be written as follows:

\begin{equation}
	P(\textbf{v})=\sum_{\textbf{h}^1}P(\textbf{h}^1)P(\textbf{v}|\textbf{h}^1),
\end{equation}
where $P(\textbf{h}^1)=\sum_{\textbf{v}}P(\textbf{h}^1,\textbf{v})$. Further, we shall proceed with the learning process of the second RBM, which then replaces $P(\textbf{h}^1)$ by 
$P(\textbf{h}^1)=\sum_{\textbf{h}^2}P(\textbf{h}^1,\textbf{h}^2)$. Roughly speaking, using such procedure, the conditional probabilities given by Equations~\ref{e.probv_dbm}-\ref{e.probh1_dbm}, and 
Contrastive Divergence, one can learn DBM parameters one layer at a time~\cite{Salakhutdinov:12}.

\subsection{Temperature-based Deep Boltzmann Machines}
\label{ss.tdbm}

Li et. al.~\cite{li2016temperature} showed that a temperature parameter $T$ controls the sharpness of the logistic-sigmoid function. In order to incorporate the temperature effect into the RBM context, they introduced this parameter to the joint distribution of the vectors $\textbf{v}$ and $\textbf{h}$ in Equation~\ref{e.probability_configuration}, which can be rewritten as follows:
\begin{equation}
\label{e.joint_distribution_with_temperature}
P(\textbf{v},\textbf{h},T)=\frac{1}{Z}e^{\frac{-E(\textbf{v},\textbf{h})}{T}}.
\end{equation}
When $T=1$, the aforementioned equation degenerates to Equation~\ref{e.probability_configuration}. In addition, Equation~\ref{e.probh} can be rewritten in order to accommodate the temperature parameter as follows:

\begin{equation}
\label{e.probh_T}
P(h_j=1|\textbf{v})=\phi\left(\frac{\sum_{i=1}^mw_{ij}v_i}{T}\right).
\end{equation}
Notice the temperature parameter does not affect the conditional probability of the input units (Equation~\ref{e.probv}).

In order to apply the very same idea to DBMs, the conditional probabilities over the two hidden layers given by Equations~\ref{e.probh2_dbm} and~\ref{e.probh1_dbm} can be derived and expressed using the following formulation, respectively:

\begin{equation}
\label{e.probh2_dbm_temperature}
	P(h^2_z=1|\textbf{h}^1)=\phi\left(\frac{\sum_{i=1}^{m^2}w^2_{iz}h^1_i}{T}\right),
\end{equation}
and 

\begin{equation}
\label{e.probh1_dbm_temperature}
	P(h^1_j=1|\textbf{v},\textbf{h}^2)=\phi\left(\frac{\sum_{i=1}^{m^1}w^1_{ij}v_i}{T}+\sum_{z=1}^{n^2}w^2_{jz}h^2_z\right).
\end{equation}

\section{Methodology}
\label{s.methodology}

In this section, we present the methodology employed to evaluate the proposed approach, as well the datasets and the experimental setup.

\subsection{Datasets}
\label{ss.datasets}

We propose to evaluate the behaviour of DBMs under different temperatures in the context of binary image reconstruction using three public datasets, as described below:

\begin{itemize}
\item MNIST dataset\footnote{\url{http://yann.lecun.com/exdb/mnist/}}: it is composed of images of handwritten digits. The original version contains a training set with $60,000$ images from digits `0'-`9', as well as a test set with $10,000$ images\footnote{The images are originally available in grayscale with resolution of $28\times28$, but they were reduced to $14\times14$ images.}. Due to the high computational burden for DBM model selection, we decided to employ the original test set together with a reduced version of the training set\footnote{The original training set was reduced to $2\%$ of its former size, which corresponds to $1,200$ images.}.
\item CalTech 101 Silhouettes Data Set\footnote{\url{https://people.cs.umass.edu/~marlin/data.shtml}}: it is based on the former Caltech 101 dataset, and it comprises silhouettes of images from 101 classes with resolution of $28\times28$. We have used only the training and test sets, since our optimization model aims at minimizing the MSE error over the training set.
\item Semeion Handwritten Digit Data Set\footnote{\url{https://archive.ics.uci.edu/ml/datasets/Semeion+Handwritten+Digit}}: it is formed by $1,593$ images from handwritten digits `0' - `9' written in two ways: the first time in a normal way (accurately) and the second time in a fast way (no accuracy). In the end, they were stretched with resolution of $16\times16$ in a grayscale of $256$ values and then each pixel was binarized.
\end{itemize}

  
\subsection{Experimental Setup}
\label{ss.setup}

We employed a $3$-layered architecture for all datasets as follows: $i$-$500$-$500$-$2000$, where $i$ stands for the number of pixels used as input for each dataset, i.e. 196 ($14\times14$ images), 784 ($28\times28$ images) and 256 ($16\times16$ images) considering MNIST, Caltech 101 Silhouettes and Semeion Handwritten Digit datasets, respectively. Therefore, we have a first and a second hidden layers with $500$ neurons each, followed by a third hidden layer with $2000$ neurons. Since this architecture has been commonly employed in several works in the literature, we opted to employ it in our work either. The remaining parameters used during the learning steps were chosen empirically and fixed for each layer as follows: $\eta = 0.1$ (learning rate), $\lambda = 0.1$ (weight decay), $\alpha = 0.00001$ (penalty parameter). In addition, we compared DBMs against DBNs using the very same configuration, i.e. architecture and parameters.

\begin{sloppypar}
In order to provide a statistical analysis by means of the Wilcoxon signed-rank 
test with significance of $0.05$~\cite{Wilcoxon:45}, we conducted a cross-validation procedure with $20$ runnings. In regard to the temperature, we considered a set of values within the range $T\in\{0.1, 0.2, 0.5, 0.8, 1.0, 1.2, 1.5, 2.0\}$ for the sake of comparison purposes. 
\end{sloppypar}

\begin{sloppypar}
Finally, we employed $30$ epochs for DBM and DBN learning weights procedure with mini-batches of size $20$. In order to provide a more precise experimental validation, we trained both DBMs and DBNs with two different algorithms\footnote{One 
sampling iteration was used for all learning algorithms.}: Contrastive 
Divergence (CD)~\cite{Hinton:02} and Persistent Contrastive Divergence 
(PCD)~\cite{Tieleman:08}.\newline
\end{sloppypar}

\section{Experimental Results}
\label{s.experiments}

In this section, we present the experimental results concerning the proposed temperature-based Deep Boltzmann Machine over three public datasets aiming at the task of binary image reconstruction. Table I presents the results considering Semeion Handwritten Digit dataset, in which the values in bold stand for the most accurate ones by means of the Wilcoxon signed-rank test. Notice we considered the minimum squared error (MSE) over the test set as the measure for comparison purposes. 

\begin{table}[!htb]
\label{t.semeion}
\begin{center}
\caption{Average MSE over the test set considering Semeion Handwritten Digit dataset.}
\resizebox{\columnwidth}{!}{%
\begin{tabular}{l|l|l|l|l|l|l|l|l|}
\cline{2-9}
                          & 0.1    & 0.2    & 0.5   & 0.8    & 1.0    & 1.2   & 1.5    & 2.0    \\ \hline
\multicolumn{1}{|c|}{\cellcolor[HTML]{D2D2D2}DBM-CD}  &  \cellcolor[HTML]{EFEFEF} 0.19919    &  \cellcolor[HTML]{EFEFEF} 0.20060     &  \cellcolor[HTML]{EFEFEF} \textbf{0.19483}     &  \cellcolor[HTML]{EFEFEF}  \textbf{0.19525}   &   \cellcolor[HTML]{EFEFEF} \textbf{0.19577}    &   \cellcolor[HTML]{EFEFEF} 0.20250   &   \cellcolor[HTML]{EFEFEF} 0.21428    &   \cellcolor[HTML]{EFEFEF} 0.23187   \\ \hline
\multicolumn{1}{|c|}{\cellcolor[HTML]{D2D2D2}DBM-PCD}  &  \cellcolor[HTML]{EFEFEF} 0.19819    &  \cellcolor[HTML]{EFEFEF} 0.19986     &  \cellcolor[HTML]{EFEFEF} \textbf{0.19506}     &  \cellcolor[HTML]{EFEFEF} \textbf{0.19424}   &   \cellcolor[HTML]{EFEFEF} 0.19698    &   \cellcolor[HTML]{EFEFEF} 0.20226   &   \cellcolor[HTML]{EFEFEF} 0.21482    &   \cellcolor[HTML]{EFEFEF} 0.23165   \\ \hline
\multicolumn{1}{|c|}{\cellcolor[HTML]{D2D2D2}DBN-CD}  &  \cellcolor[HTML]{EFEFEF} 0.21613    &  \cellcolor[HTML]{EFEFEF} 0.21977     &  \cellcolor[HTML]{EFEFEF} 0.21814     &  \cellcolor[HTML]{EFEFEF}  0.21465   &   \cellcolor[HTML]{EFEFEF} 0.21352    &   \cellcolor[HTML]{EFEFEF} 0.21413   &   \cellcolor[HTML]{EFEFEF} 0.21725    &   \cellcolor[HTML]{EFEFEF} 0.22455   \\ \hline
\multicolumn{1}{|c|}{\cellcolor[HTML]{D2D2D2}DBN-PCD}  &  \cellcolor[HTML]{EFEFEF} 0.21051    &  \cellcolor[HTML]{EFEFEF} 0.21155     &  \cellcolor[HTML]{EFEFEF} 0.21660     &  \cellcolor[HTML]{EFEFEF}  0.21104   &   \cellcolor[HTML]{EFEFEF} 0.21012    &   \cellcolor[HTML]{EFEFEF} 0.21031   &   \cellcolor[HTML]{EFEFEF} 0.21080    &   \cellcolor[HTML]{EFEFEF} 0.21431   \\ \hline
\end{tabular}}
\end{center}
\end{table}

Clearly, one can not observe a statistical difference between DBM-CD and DBM-PCD when using $T\in\{0.5,0.8,1.0\}$, except for DBM-PCD and $T=1$. Also, our results confirm the ones obtained by Lin et al.~\cite{li2016temperature}, i.e. the lower the temperature the higher the entropy. In short, we can learn more information at low temperatures, thus obtaining better results (obviously, we are constrained to a minimum bound concerning the temperature). According to Ranzato et al.~\cite{RanzatoNIPS:08}, sparsity in the neuron's activity favours the power of generalization of a network, which is somehow related to dropping neurons out in order to avoid overfitting~\cite{Srivastava:14}. We have observed the following statement: the lower the temperature, the higher the probability of turning ``on"\ hidden units (Equation~\ref{e.probh1_dbm_temperature}), which forces DBM to push down the weights ($\textbf{W}$) looking at sparsity. When we push the weights down, we also decrease the probability of turning on the hidden units, i.e. we try to deactivate them, thus forcing the network to learn by other ways. We observed the process of pushing the weights down to be more ``radical"\ at lower temperatures.

\begin{figure}[ht]
  \centerline{
    \begin{tabular}{cccc}
	\includegraphics[width=2.67cm,height=2.67cm]{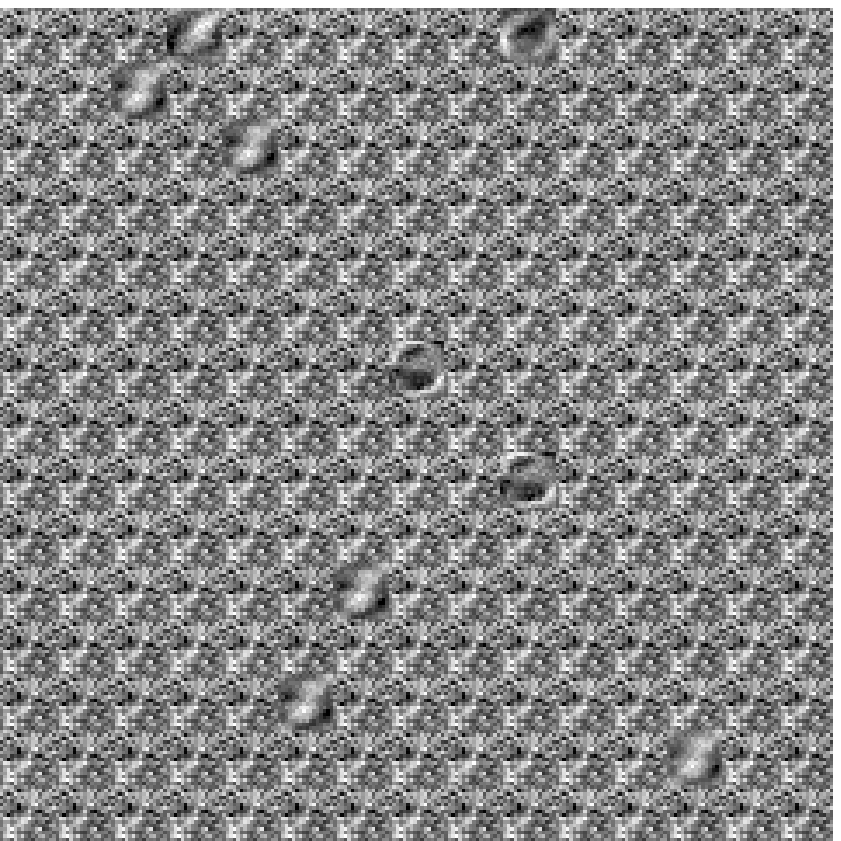} &
	\includegraphics[width=2.67cm,height=2.67cm]{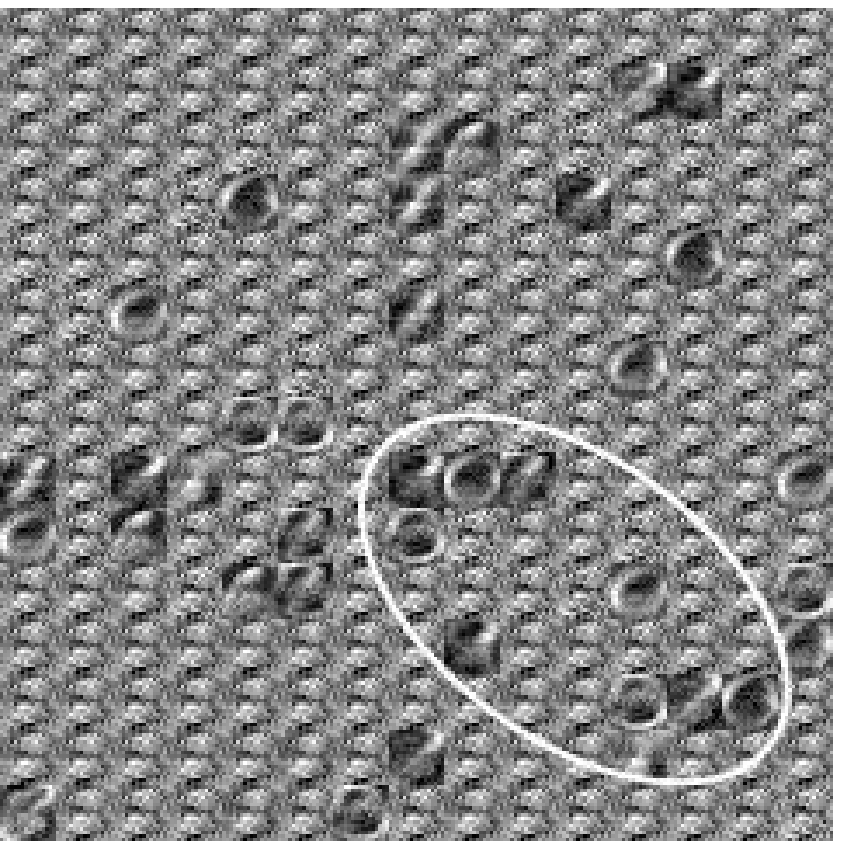} &
	\includegraphics[width=2.67cm,height=2.67cm]{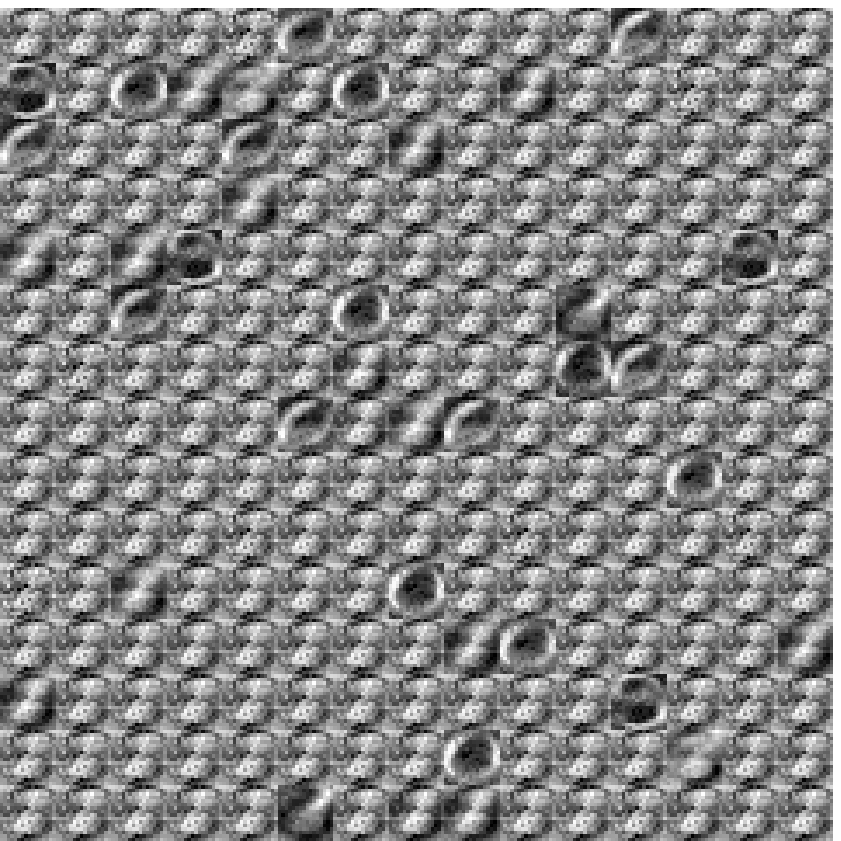} & 
	\includegraphics[width=2.67cm,height=2.67cm]{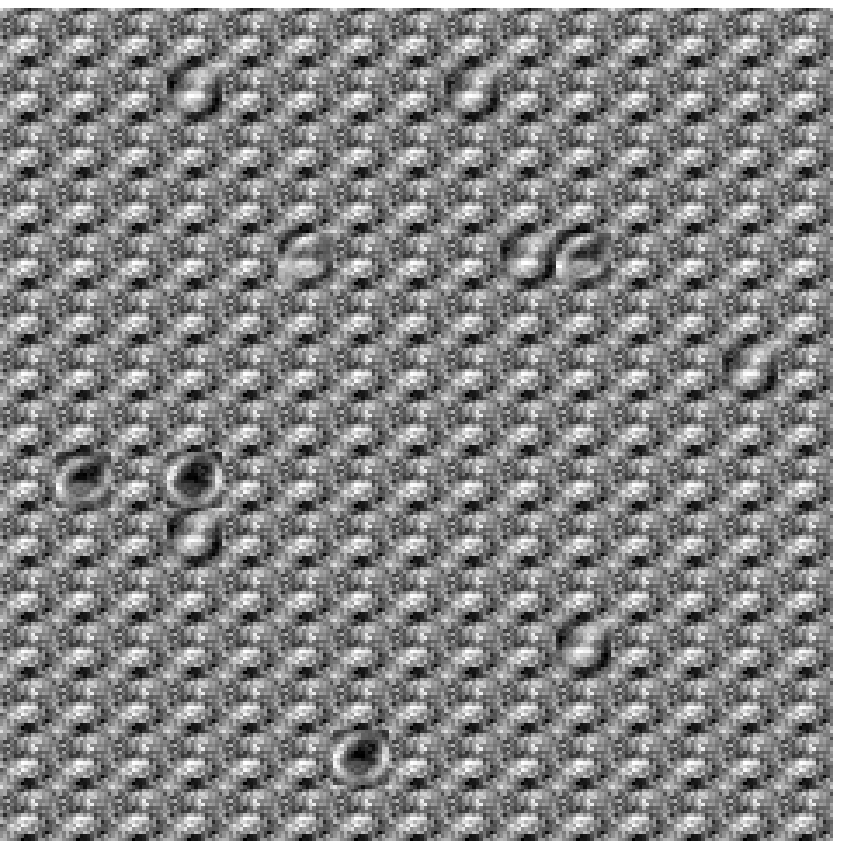} \\
	(a) & (b) & (c) & (d)
    \end{tabular}}
    \caption{Effect of different temperatures by means of DBM-PCD considering Semeion Handwritten Digit dataset with respect to the connection weights of the first hidden layer for: (a) $T=0.1$, (b) $T=0.5$, (c) $T=1.0$, (d) $T=2.0$.}
  \label{f.semeion}
  \end{figure}

Figure~\ref{f.semeion} displays the values of the connection weights between the input and the first hidden layer. Since we used an architecture with $500$ hidden neurons in the first layer, we chose $225$ neurons at random to display what sort of information they have learned. According to Table I, some of the better results were obtained using $T=0.5$ (Figure~\ref{f.semeion}b) and $T=1$ (Figure~\ref{f.semeion}c), which can be observed in the images either. Notice we can observe some digits at these images (e.g. highlighted regions in Figure~\ref{f.semeion}b), while they are scarce in others. Additionally, DBNs seemed to benefit from lower temperatures, but their results were inferior to the ones obtained by DBMs.

\begin{figure}[ht]
  \centerline{
    \begin{tabular}{cccc}
	\includegraphics[width=2.67cm,height=2.67cm]{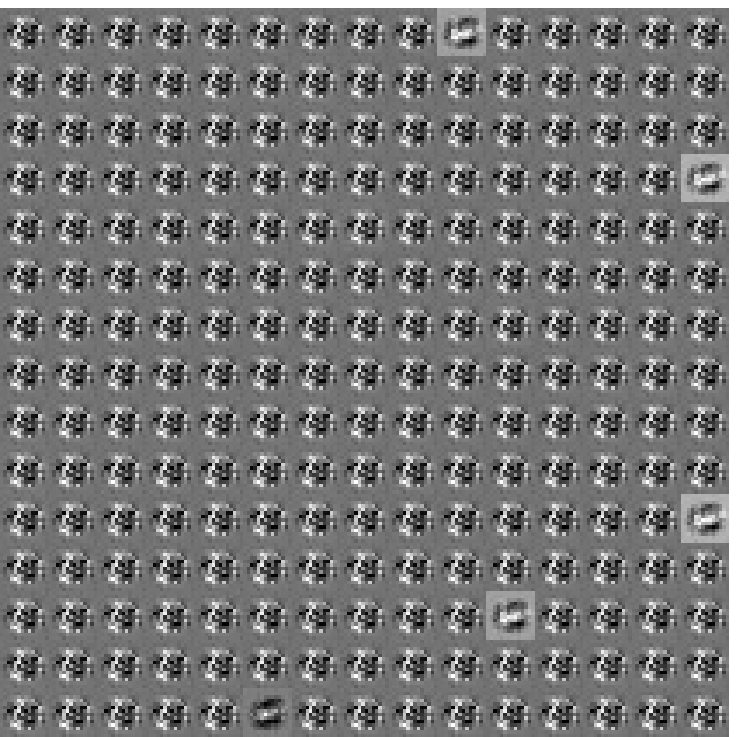} & 
	\includegraphics[width=2.67cm,height=2.67cm]{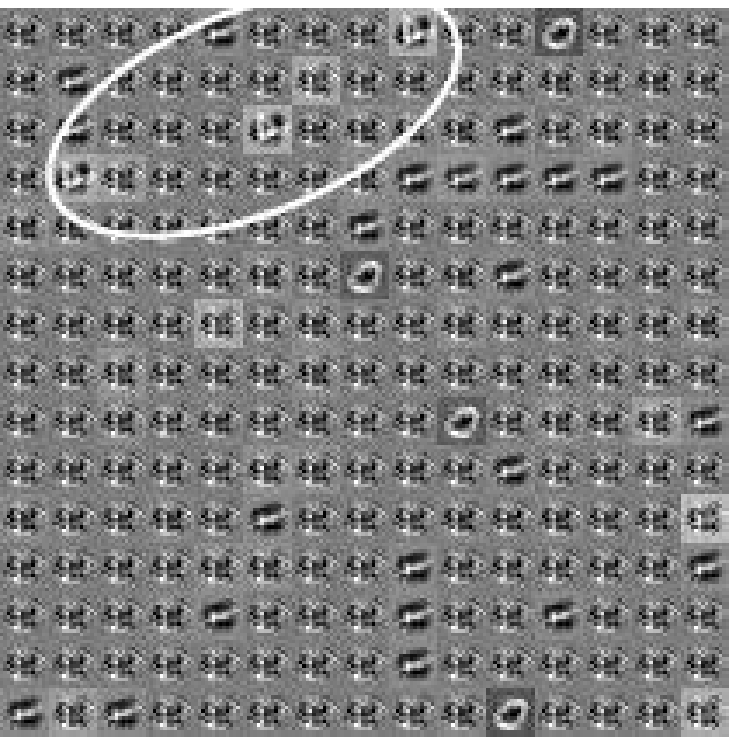} &
	\includegraphics[width=2.67cm,height=2.67cm]{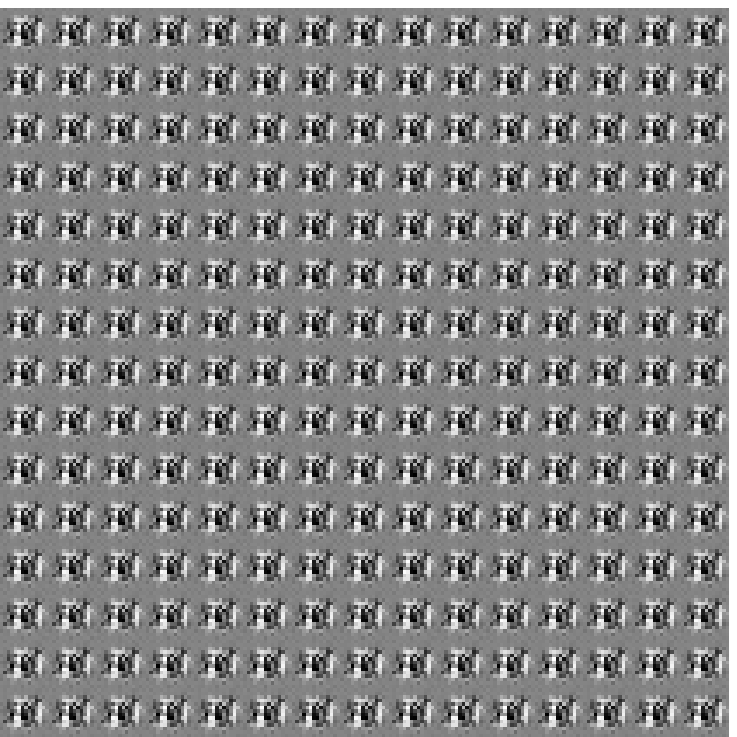} & 
	\includegraphics[width=2.67cm,height=2.67cm]{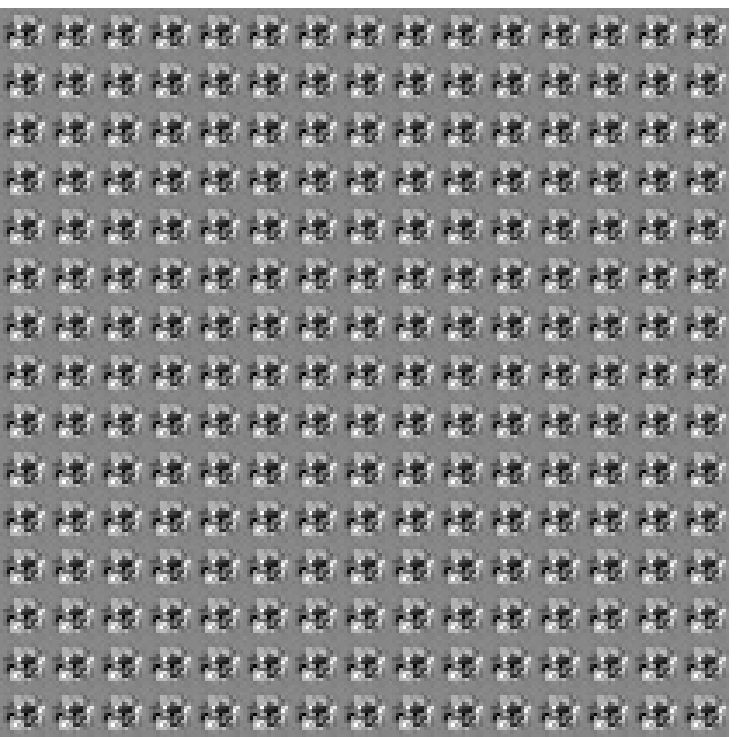} \\
	(a) & (b) & (c) & (d)
    \end{tabular}}
    \caption{Effect of different temperatures by means of DBM-PCD considering MNIST dataset with respect to the connection weights of the first hidden layer for: (a) $T=0.1$, (b) $T=0.5$, (c) $T=1.0$, (d) $T=2.0$.}
  \label{f.mnist}
  \end{figure}

Table II displays the MSE results over MNIST dataset, where the best results were obtained with $T=0.1$. Once again, the results confirmed the hypothesis that better results can be obtained at lower temperatures, probably due to the lower interaction between visible and hidden units, which may imply in a slower convergence, but avoiding local optima (learning in DBMs is essentially an optimization problem, where we aim at minimizing the energy of each training sample in order to increase its probability - Equations~\ref{e.dbm_energy} and~\ref{e.probability_sample_dbm}). Figure~\ref{f.mnist} displays the connection weights between the input and the first hidden layer concerning DBM-PCD, where the highlighted region depicts some important information learned from the hidden neurons. Notice the neurons do not seem to contribute a lot with respect to different information learned from each other at higher temperatures (Figure~\ref{f.mnist}d), since most of them have similar information encoded.

\begin{table}[!htb]
\begin{center}
\caption{Average MSE over the test set considering MNIST dataset.}
\resizebox{\columnwidth}{!}{%
\begin{tabular}{l|l|l|l|l|l|l|l|l|}
\cline{2-9}
                          & 0.1    & 0.2    & 0.5   & 0.8    & 1.0    & 1.2   & 1.5    & 2.0    \\ \hline
\multicolumn{1}{|c|}{\cellcolor[HTML]{D2D2D2}DBM-CD}  &  \cellcolor[HTML]{EFEFEF} \textbf{0.08647}    &  \cellcolor[HTML]{EFEFEF} 0.08768     &  \cellcolor[HTML]{EFEFEF} 0.08786     &  \cellcolor[HTML]{EFEFEF}  0.08881   &   \cellcolor[HTML]{EFEFEF} 0.09111    &   \cellcolor[HTML]{EFEFEF} 0.09222   &   \cellcolor[HTML]{EFEFEF} 0.09402    &   \cellcolor[HTML]{EFEFEF} 0.09629   \\ \hline
\multicolumn{1}{|c|}{\cellcolor[HTML]{D2D2D2}DBM-PCD}  &  \cellcolor[HTML]{EFEFEF} \textbf{0.08637}    &  \cellcolor[HTML]{EFEFEF} 0.08759     &  \cellcolor[HTML]{EFEFEF} 0.08793     &  \cellcolor[HTML]{EFEFEF}  0.08870   &   \cellcolor[HTML]{EFEFEF} 0.09106    &   \cellcolor[HTML]{EFEFEF} 0.09231   &   \cellcolor[HTML]{EFEFEF} 0.09405    &   \cellcolor[HTML]{EFEFEF} 0.09618   \\ \hline
\multicolumn{1}{|c|}{\cellcolor[HTML]{D2D2D2}DBN-CD}  &  \cellcolor[HTML]{EFEFEF} 0.08993    &  \cellcolor[HTML]{EFEFEF} 0.09432     &  \cellcolor[HTML]{EFEFEF} 0.09259     &  \cellcolor[HTML]{EFEFEF} 0.09012   &   \cellcolor[HTML]{EFEFEF} 0.08933    &   \cellcolor[HTML]{EFEFEF} 0.08924   &   \cellcolor[HTML]{EFEFEF} 0.08966    &   \cellcolor[HTML]{EFEFEF} 0.09110   \\ \hline
\multicolumn{1}{|c|}{\cellcolor[HTML]{D2D2D2}DBN-PCD}  &  \cellcolor[HTML]{EFEFEF} 0.08784    &  \cellcolor[HTML]{EFEFEF} 0.08811     &  \cellcolor[HTML]{EFEFEF} 0.08919     &  \cellcolor[HTML]{EFEFEF}  0.08874   &   \cellcolor[HTML]{EFEFEF} 0.08833    &   \cellcolor[HTML]{EFEFEF} 0.08820   &   \cellcolor[HTML]{EFEFEF} 0.08838    &   \cellcolor[HTML]{EFEFEF} 0.08994   \\ \hline
\end{tabular}}
\end{center}
\label{t.mnist}
\end{table}

\begin{figure}[ht]
  \centerline{
    \begin{tabular}{cccc}
	\includegraphics[width=2.67cm,height=2.67cm]{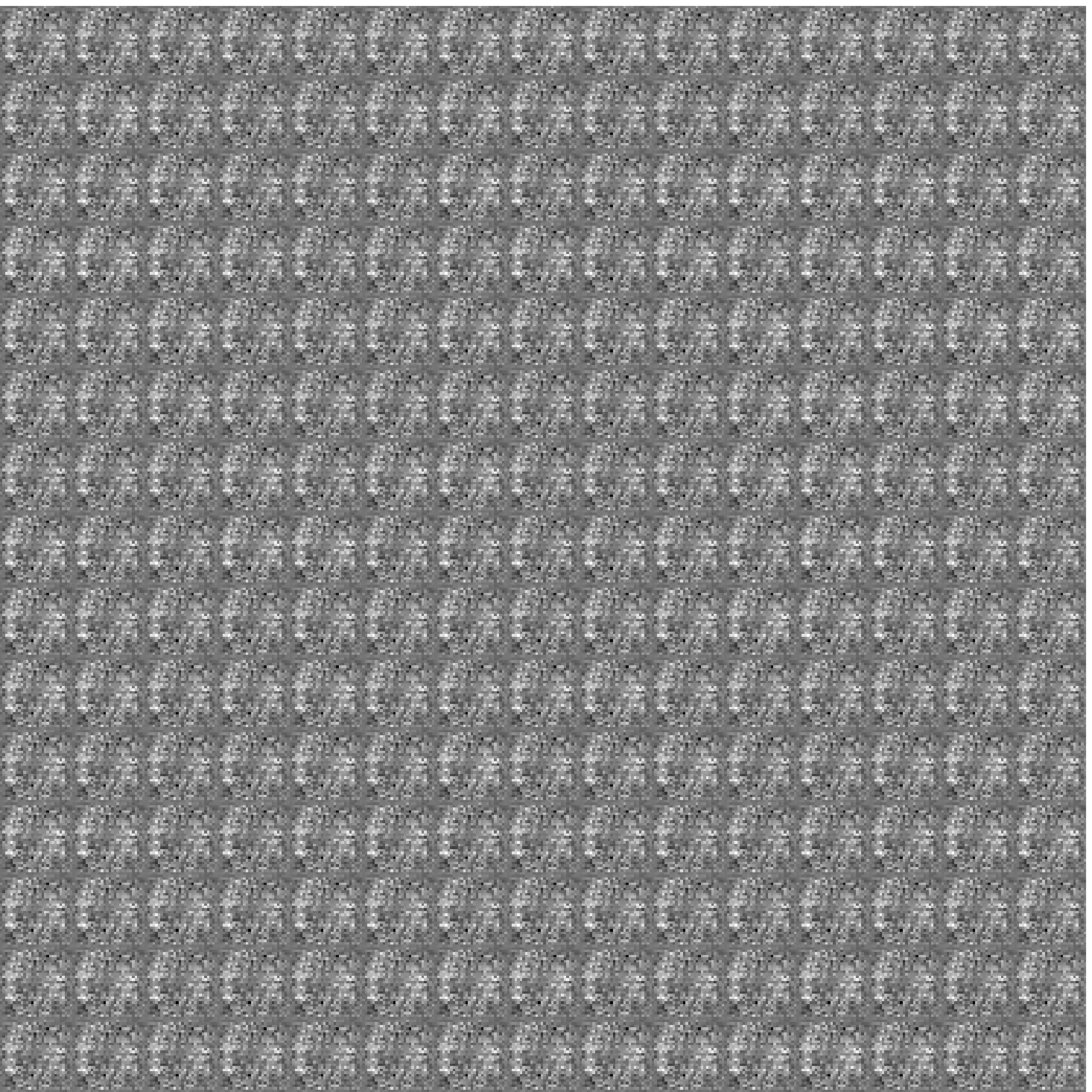} & 
	\includegraphics[width=2.67cm,height=2.67cm]{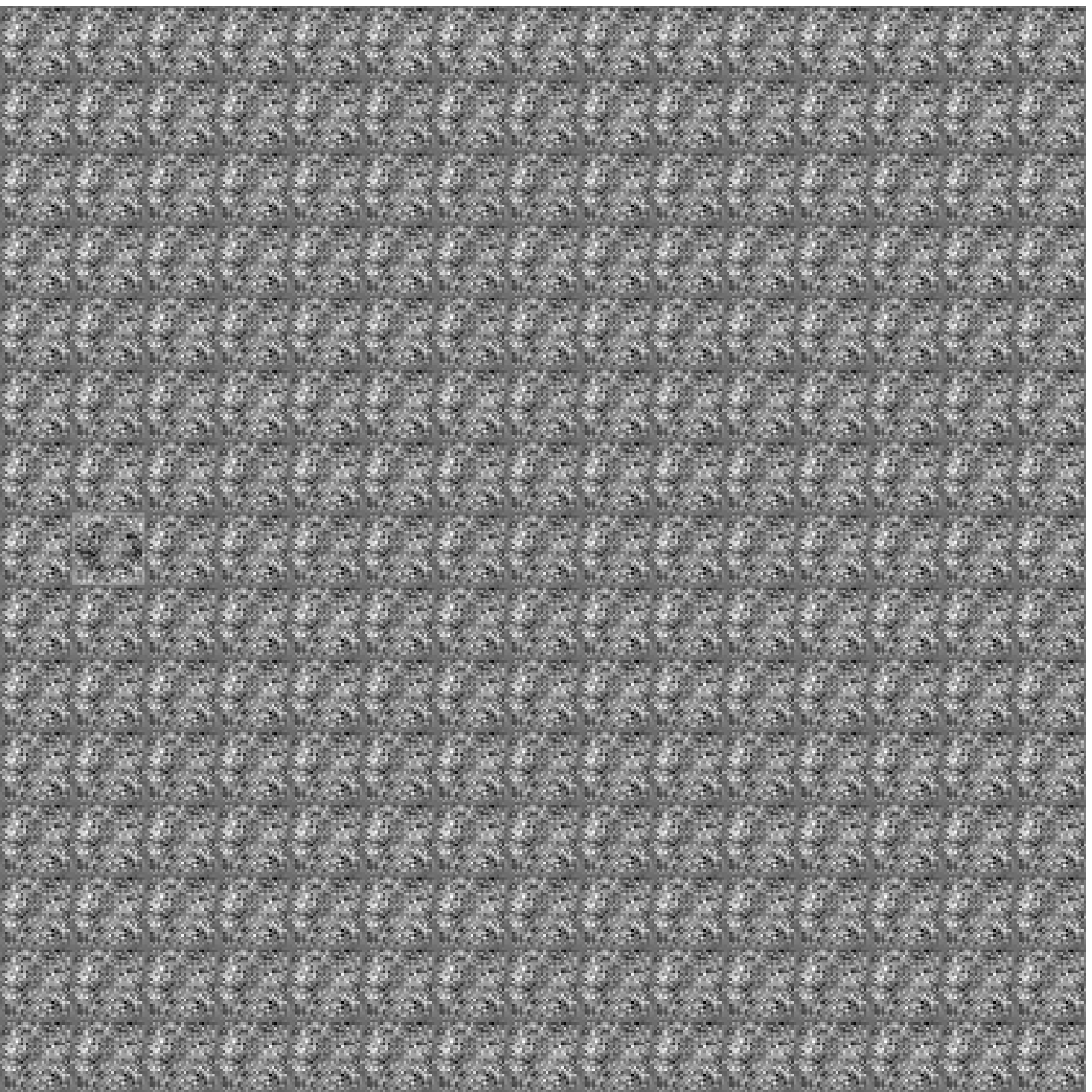} &
	\includegraphics[width=2.67cm,height=2.67cm]{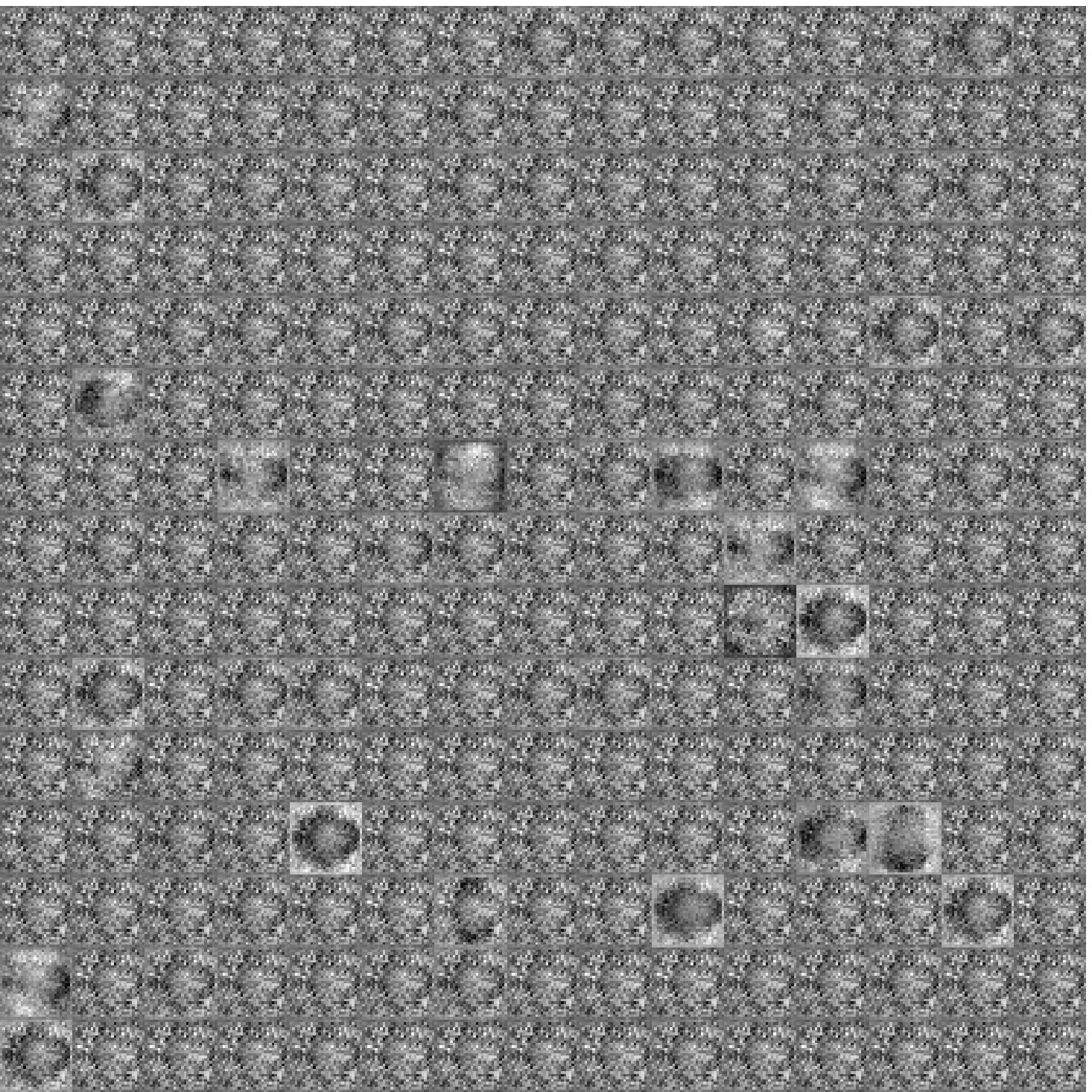} & 
	\includegraphics[width=2.67cm,height=2.67cm]{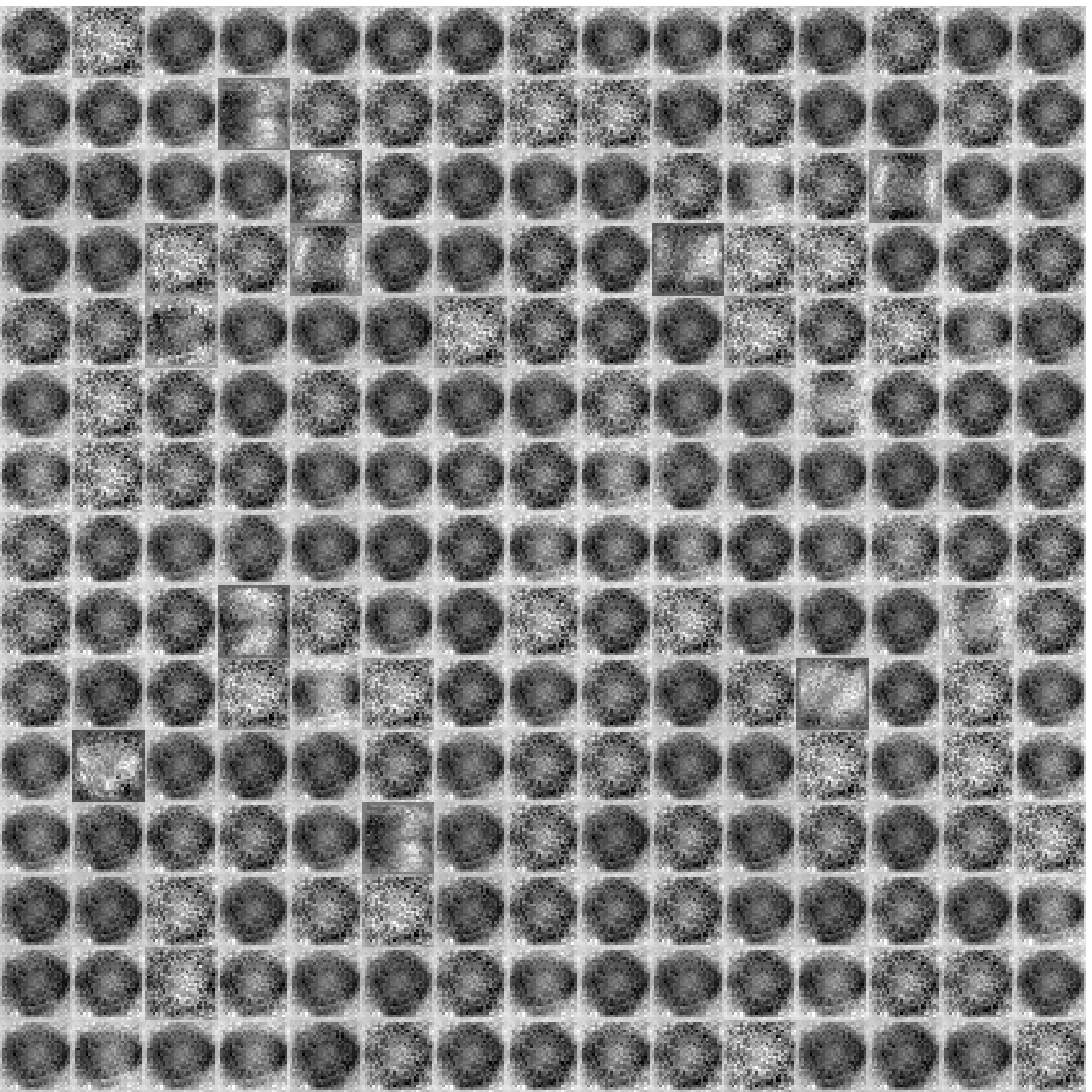} \\
	(a) & (b) & (c) & (d)
    \end{tabular}}
    \caption{Effect of different temperatures by means of DBM-PCD considering Caltech 101 Silhouettes dataset with respect to the connection weights of the first hidden layer for: (a) $T=0.1$, (b) $T=0.5$, (c) $T=1.0$, (d) $T=2.0$..}
  \label{f.caltech}
  \end{figure}

Table III presents the MSE results obtained over Caltech 101 Silhouettes dataset, where the lower temperatures obtained the best results. In this case, both DBM and DBN obtained similar results. Since this dataset comprises a number of different objects and classes, it is more complicated to figure out some shape with respect to the neurons' activity in Figure~\ref{f.caltech}. Curiously, the neurons' response at the lower temperatures (Figure~\ref{f.caltech}a) led to a different behaviour that has been observed in the previous datasets, since the more ``active"\ neurons with respect to different information learned were the ones obtained with $T=2$ at the training step. We believe such behaviour is due to the number of iterations for learning used in this paper, which might not be enough for convergence purposes at lower temperatures, since this dataset poses a greater challenge than the others (it has a great intra-class variablity).

\begin{table}[!htb]
\begin{center}
\caption{Average MSE over the test set considering Caltech 101 Silhouettes dataset.}
\resizebox{\columnwidth}{!}{%
\begin{tabular}{l|l|l|l|l|l|l|l|l|}
\cline{2-9}
                          & 0.1    & 0.2    & 0.5   & 0.8    & 1.0    & 1.2   & 1.5    & 2.0    \\ \hline
\multicolumn{1}{|c|}{\cellcolor[HTML]{D2D2D2}DBM-CD}  &  \cellcolor[HTML]{EFEFEF} \textbf{0.16072}    &  \cellcolor[HTML]{EFEFEF} 0.16119     &  \cellcolor[HTML]{EFEFEF} 0.16304     &  \cellcolor[HTML]{EFEFEF}  0.16335   &   \cellcolor[HTML]{EFEFEF} 0.16335    &   \cellcolor[HTML]{EFEFEF} 0.16375   &   \cellcolor[HTML]{EFEFEF} 0.16410    &   \cellcolor[HTML]{EFEFEF} 0.16390   \\ \hline
\multicolumn{1}{|c|}{\cellcolor[HTML]{D2D2D2}DBM-PCD}  &  \cellcolor[HTML]{EFEFEF} \textbf{0.16068}    &  \cellcolor[HTML]{EFEFEF} 0.16125     &  \cellcolor[HTML]{EFEFEF} 0.16295     &  \cellcolor[HTML]{EFEFEF}  0.16387   &   \cellcolor[HTML]{EFEFEF} 0.16359    &   \cellcolor[HTML]{EFEFEF} 0.16451   &   \cellcolor[HTML]{EFEFEF} 0.16368    &   \cellcolor[HTML]{EFEFEF} 0.16389   \\ \hline
\multicolumn{1}{|c|}{\cellcolor[HTML]{D2D2D2}DBN-CD}  &  \cellcolor[HTML]{EFEFEF} \textbf{0.16061}    &  \cellcolor[HTML]{EFEFEF} 0.16107     &  \cellcolor[HTML]{EFEFEF} 0.16269     &  \cellcolor[HTML]{EFEFEF}  0.16301   &   \cellcolor[HTML]{EFEFEF} 0.16320    &   \cellcolor[HTML]{EFEFEF} 0.16310   &   \cellcolor[HTML]{EFEFEF} 0.16320    &   \cellcolor[HTML]{EFEFEF} 0.16282   \\ \hline
\multicolumn{1}{|c|}{\cellcolor[HTML]{D2D2D2}DBN-PCD}  &  \cellcolor[HTML]{EFEFEF} \textbf{0.16062}    &  \cellcolor[HTML]{EFEFEF} 0.16085     &  \cellcolor[HTML]{EFEFEF} 0.16146     &  \cellcolor[HTML]{EFEFEF}  0.16158   &   \cellcolor[HTML]{EFEFEF} 0.16158    &   \cellcolor[HTML]{EFEFEF} 0.16190   &   \cellcolor[HTML]{EFEFEF} 0.16176    &   \cellcolor[HTML]{EFEFEF} 0.16267   \\ \hline
\end{tabular}}
\\~\\
\end{center}
\label{t.tableCaltech}
\end{table}

\section{Conclusions and Future Works}
\label{s.conclusions}

In this work, we dealt with the problem of different temperatures at the DBM learning step. Inspired by a very recent work that proposed the Temperature-based Restricted Boltzmann Machines~\cite{li2016temperature}, we decided to evaluate the influence of the temperature when learning with Deep Boltzmann Machines aiming at the task of binary image reconstruction. Our results confirm the hypothesis raised by Li et al.~\cite{li2016temperature}, where the lower the temperature, the more generalized is the network. Thus, more accurate results can be obtained.

We observed the network pushes the weights down at lower temperatures in order to favour the sparsity, since the probability of tuning on hidden units is greater at lower temperatures. In regard to future works, we aim to propose an adaptive temperature, which can be linearly increased/decreased along the iterations in order to speed up the convergence process.

\section*{Acknowledgments}
The authors would like to thank FAPESP grant \#2014/16250-9, Capes and CNPq grant \#306166/2014-3.

\bibliographystyle{spmpsci}
\bibliography{paper}

\end{document}